\documentclass{article}

\usepackage[preprint]{neurips_2022}

\usepackage[utf8]{inputenc} % allow utf-8 input
\usepackage[T1]{fontenc}    % use 8-bit T1 fonts
\usepackage{hyperref}       % hyperlinks
\usepackage{url}            % simple URL typesetting
\usepackage{booktabs}       % professional-quality tables
\usepackage{amsfonts}       % blackboard math symbols
\usepackage{nicefrac}       % compact symbols for 1/2, etc.
\usepackage{microtype}      % microtypography
\usepackage{xcolor}         % colors

\usepackage{subcaption}
\usepackage{algorithm}
\usepackage[noend]{algpseudocode}
\usepackage{graphicx}
\usepackage{amsmath}

\title{Adaptive Rollout Length for Model-Based RL Using Model-Free Deep RL}

\author{%
  Abhinav Bhatia \qquad~~ Philip S. Thomas \qquad~~ Shlomo Zilberstein \\
  College of Information and Computer Sciences\\
  University of Massachusetts Amherst, MA 01003 \\
  \texttt{\{abhinavbhati,pthomas,shlomo\}@cs.umass.edu}
}

\newcommand{\argmax}{\text{argmax}}

\begin{document}

\maketitle

\begin{abstract}
    Model-based reinforcement learning promises to learn an optimal policy from fewer interactions with the environment compared to model-free reinforcement learning by learning an intermediate model of the environment in order to predict future interactions. When predicting a sequence of interactions, the \emph{rollout length}, which limits the prediction horizon, is a critical hyperparameter as accuracy of the predictions diminishes in the regions that are further away from real experience. As a result, with a longer rollout length, an overall worse policy is learned in the long run. Thus, the hyperparameter provides a trade-off between quality and efficiency.
    
    In this work, we frame the problem of tuning the rollout length as a meta-level sequential decision-making problem that optimizes the final policy learned by model-based reinforcement learning given a fixed budget of environment interactions by adapting the hyperparameter dynamically based on feedback from the learning process, such as accuracy of the model and the remaining budget of interactions. We use model-free deep reinforcement learning to solve the meta-level decision problem and demonstrate that our approach outperforms common heuristic baselines on two well-known reinforcement learning environments.
\end{abstract}

\section{Introduction}

Reinforcement learning algorithms fall into two broad categories---model-based and model-free---depending on whether or not they construct an intermediate model. Model-free reinforcement learning methods based on value-function approximation have been successfully applied to a wide array of domains such as playing video games from raw pixels~\citep{mnih2015human} and motor control tasks~\citep{lillicrap2015continuous,haarnoja2018soft}. However, model-free methods require a large number of interactions with the environment to compensate for the unknown dynamics and limited generalizability of the value-function. In contrast, model-based approaches promise better efficiency by constructing a model of the environment from the interactions, and using it to guess the outcomes of future interactions. Therefore, model-based methods are better suited for learning in the real world, where environment-interactions are expensive and often limited. 

However, using the model is not straightforward and requires considering multiple factors, such as how the model is learned, how accurate it is, how well it generalizes, and whether it is optimistic or pessimistic. When the model is queried to predict a future trajectory, the look-ahead horizon, or the rollout length, limits \emph{how much} the model is used by limiting how further away from real experience the model operates. Intuitively, a longer horizon permits greater efficiency by making greater use of the model. However, when the model is asked to predict a long trajectory, the predictions get less accurate at each step as the errors compound, leading to worse performance in the long run. This suggests that the rollout length should be based at least on the accuracy of the model and the budget of available interactions with the environment. 

Given the importance of this hyperparameter, it is surprising that there is little prior work that adjusts the rollout length dynamically in a manner that is aware of certain important evolving features of the learning process. In this work, we frame the problem of adjusting the rollout length as a meta-level closed-loop sequential decision making problem---a form of metareasoning aimed at achieving bounded rationality~\citep{russell1991principles,SZ:Zmetareasoning11}. Our meta-level objective is to arrive at a rollout length adjustment strategy that optimizes the final policy learned by the model-based reinforcement learning agent given a bounded budget of environment interactions. We solve the meta-level control problem using model-free deep reinforcement learning, an approach that has been proposed for dynamic hyperparameter tuning in general~\citep{biedenkapp2020dynamic} and has been shown to be effective for solving decision-theoretic meta-level control problems in anytime planning~\citep{bhatia2022tuning}. In our case, we train a model-free deep reinforcement learning metareasoner on many instances of model-based reinforcement learning applied to simplified simulations of the target real-world environment, so that the trained metareasoner can be transferred to control the rollout length for model-based reinforcement learning in the real world.

We experiment with \textsc{Dqn}~\citep{mnih2015human} model-free reinforcement learning algorithm as a metareasoner for adjusting the rollout length for \textsc{Dyna-Dqn}~\citep{DBLP:journals/corr/abs-1806-01825} model-based reinforcement learning algorithm on classic control environments \textsc{MountainCar}~\citep{moore1990efficient} and \textsc{Acrobot}~\citep{sutton1996generalization}. We test the trained metareasoner on environments constructed by perturbing the parameters of the original environments in order to capture the gap between the real world and the simulation. The results show that our approach outperforms common approaches that adjust the rollout length using heuristic schedules.

The paper is organized as follows. First, we cover the background material relevant to our work in section 2. Next, we motivate the importance of adjusting the rollout length in a principled manner in section 3. We propose our metareasoning approach in section 4. We mention related work in section 5. We present our experiments, results and discussion in sections 6-7. Finally, we conclude the paper in section 8.

\section{Background}

\subsection{MDP Optimization Problem}

A Markov Decision Process~\citep{DBLP:books/wi/Puterman94}, or an MDP, is repsented by a tuple $<\mathcal S,
\mathcal A,p,r,d_0,\gamma>$, where $\mathcal S$ is the state space, $\mathcal A$ is the set of available actions, $p : S \times A \times S \rightarrow [0,1]$ is the transition function representing the probability $p(s,a,s')$ of transitioning from state $s$ to state $s'$ by taking action $a$. The reward function $r : \mathcal S \times \mathcal A \rightarrow \mathbb R$ represents the expected reward $r(s,a)$ associated with action $a$ in state $s$. $d_0 : \mathcal S \rightarrow [0,1]$ represents the probability $d_0(s)$ that the MDP process starts with state $s$. $\gamma \in [0, 1]$ denotes the discount factor.

An agent may act in the MDP environment according to a Markovian policy $\pi: \mathcal S \times \mathcal A \rightarrow [0,1]$, which represents the probability $\pi(s,a)$ of taking action $a$ in a state $s$. The objective is to find an optimal policy $\pi^*$ that maximizes the expected return $J(\pi)$:

\begin{align}
    J(\pi) &:= \mathbb E[\sum_{t=0}^{T-1} \gamma^t R_t | \pi] \\
    \pi^* &:= \underset{\pi \in \Pi}{\argmax} J(\pi)
\end{align}

where random variable $R_t$ denotes the reward obtained at timestep $t$ and random variable $T$ denotes the number of steps until termination of the process or an episode.

An action-value function $q^\pi(s,a)$ is defined as the expected return obtained by taking action $a$ at state $s$ and thereafter executing policy $\pi$:

\begin{align}
    q^\pi(s,a) := \mathbb E[\sum_{k=t}^{T-1} \gamma^k R_k | S_t=s, A_t=a,\pi]
\end{align}

This can be expressed recursively as the Bellman equation for the action-value function:

\begin{align}
    q^\pi(s,a) = r(s,a) + \gamma \sum_{s'} \sum_{a'} p(s,a,s') \pi(s',a') q^\pi(s',a')
\end{align}

The action-value function for an optimal policy satisfies the Bellman optimality equation:
\begin{align}
    q^*(s,a) := r(s,a) + \gamma \sum_{s'} p(s,a,s') \max_{a'} q^*(s',a')
\end{align}

For every MDP, there exists an optimal deterministic policy~\citep{DBLP:books/lib/Bertsekas05, DBLP:series/synthesis/2012Mausam}, often denoted simply as the mapping $\pi^*: \mathcal S \rightarrow \mathcal A$, which can be recovered from the optimal action-value function by taking a greedy action at every state i.e., $\pi^*(s) \in \argmax_{a}q^*(s,a)$.

\subsection{Model-Free Reinforcement Learning} \label{sec:MFRL}

In reinforcement learning (RL) setting, either the transition model or the reward model or both are unknown, and the agent must derive an optimal policy by interacting with the environment and observing feedback~\citep{sutton2018reinforcement}.

Most model-free approaches to reinforcement learning maintain an estimate of the value function. When an agent takes an action $a$ at state $s$ and observes a reward $r$ and transitions to state $s'$, the estimated $q$-value of a policy $\pi$, denoted as $\hat q^\pi$, is updated using a temporal-difference update rule:
\begin{align}
    \hat q^\pi(s,a) &\gets \hat q^\pi(s,a) + \alpha (r + \gamma \mathbb E_{a'\sim \pi(s',\cdot)}[\hat q^\pi(s',a')] - \hat q^\pi(s,a)) \label{eq:td-update}
\end{align}

Where $\alpha \in [0,1]$ is the step size. Repeated application of this rule in arbitrary order despite bootstrapping from random estimates causes convergence to the true action-value function $q^\pi$, as long as the policy used to collect the experience tuples $(s,a,r,s')$ explores every state-action pair sufficiently often. Maintaining an estimate of the value function helps the agent improve its policy between value function updates by assigning greater probability mass to actions with higher $q$-values. With greedy improvements $\pi(s) \gets \argmax_a \hat q^\pi(s,a)$, the process converges to the optimal action-value function $q*$, an approach known as \textsc{q}-learning~\citep{watkins1992q}. A common choice for the exploration policy is an $\epsilon$-greedy policy -- which selects an action randomly with probability $\epsilon$ when not selecting a greedy action.

When the state space is continuous, the value function may be parameterized using a function approximator with parameters $\theta$ to allow generalization to unexplored states. The popular algorithm Deep-Q-Network, or \textsc{Dqn}, uses deep learning to approximate the action-value function~\citep{mnih2015human}. A gradient update of \textsc{Dqn} minimizes the following loss.

\begin{align}
    \mathcal L(\theta) = \underset{(s,a,r,s') \sim \mathcal D}{\mathbb E} [(r + \gamma \max_{a'} q_{\theta'}(s',a') - q_\theta(s,a))^2]
\end{align}

Where the parameters $\theta'$ are assigned $\theta' \gets \theta$ over spaced-out intervals to stabilize the loss function. A minibatch of experience tuples $(s,a,r,s')$ is sampled from an experience memory buffer $\mathcal D$ populated by acting in the environment by following an exploration policy. Reusing, or \emph{replaying}, recorded experience boosts sample efficiency by eliminating the need to revisit those transitions~\citep{DBLP:journals/ml/Lin92}.

Since \textsc{q}-learning does not require learning an intermediate  model to learn an optimal policy, it belongs to the paradigm of model-free reinforcement learning.

\subsection{Model-Based Reinforcement Learning} \label{sec:MBRL}

In model-based reinforcement learning, or MBRL, the agent learns an intermediate model from the data collected while interacting with the environment, and uses the model to derive an optimal plan.

In this work, we focus on a class of methods in which the agent learns a generative forward-dynamics model (i.e., the transition and the reward function), and uses it to synthesize novel experience, which boosts sample efficiency as it augments the experience used to learn the value function. The model can be advantageous for additional reasons, such as enabling better exploration~\citep{thrun1992}, transfer~\citep{JMLR:v10:taylor09a}, safety~\citep{berkenkamp2017safe} and explanability~\citep{moerland2018emotion}. Despite the advantages, model-based methods often converge to suboptimal policies due to the model's biases and inaccuracies.

\textsc{DynaQ}~\citep{DBLP:journals/sigart/Sutton91,DBLP:conf/icml/Sutton90} was one of the earliest approaches that integrated acting, learning and planning in a single loop in a tabular RL setting. In \textsc{DynaQ}, when the agent experiences a new transition, i) it is used to perform a standard \textsc{q}-learning update, ii) it is used to update a tabular maximum likelihood model, iii) the model is used to generate a single experience by taking an action suggested by the current policy on a randomly selected state that has been previously visited in the environment, and iv) the generated experience is used to perform a \textsc{q}-learning update.

Recent state-of-the-art approaches in model-based deep reinforcement learning have extended this architecture to perform multi-step rollouts where each transition in the synthesized trajectory is treated as an experience for model-free reinforcement learning~\citep{DBLP:journals/corr/abs-1806-01825,janner2019trust,Kaiser2020Model}.

\cite{DBLP:journals/corr/abs-2006-16712} provide a comprehensive survey of model-based reinforcement learning.

\section{Motivation}

In scenarios where the environment interactions are costly and limited, such as in the real world, model-based RL promises to be more suitable than model-free RL as it can learn a better policy from fewer interactions. However, the quality of the policy learned after a given number of environment interactions depends on many algorithm design decisions. Choosing the rollout length is a critical and a difficult decision for the following reasons.

First, we note that for model-based RL to be more efficient than model-free RL, rollouts to unfamiliar states must be accurate enough such that the subsequently derived value estimates are more accurate than the estimates obtained by replaying familiar experience and relying on value-function generalization alone. However, the model itself loses accuracy in regions further away from familiar states and moreover, prediction errors begin compounding at each step along a rollout. As a result, shorter rollouts are more accurate but provide little gain in efficiency, while longer rollouts improve efficiency in the short term but ultimately cause convergence to a suboptimal policy~\citep{DBLP:journals/corr/abs-1806-01825}. 

The ideal rollout length depends on many factors, one of which is the evolving accuracy of the learned model. For example, when the model is significantly inaccurate, which is the case in the beginning of the training, a short rollout length may be a better choice. The kind of inaccuracy -- whether optimistic or pessimistic also matters, given that planning using a pessimistic (or inadmissible) model discourages exploration. Other important factors include the quality of the policy at a given point of time during the training, and the remaining budget of environment interactions. For instance, once an agents learns a good policy with enough environment interactions to go, the agent may benifit from reducing the rollout length or even switching to model-free learning entirely in order to refine the policy using real data alone. Finally, the rollout length itself affects the policy, which affects the model's training data, and therefore affects the model.

As a result, choosing the rollout length is a complex, closed-loop, sequential decision-making problem. In other words, it requires an approach that searches for an ideal sequence of adjustments considering their long term consequences and feedback from the training process.

\section{Adjusting Rollout Length Using Metareasoning}

In this section, we describe our metareasoning approach to adjust the rollout length over the course of the training in MBRL, in order to maximize the quality of the final policy learned by the agent at the end of a fixed budget of environment interactions.

Our MBRL architecture is outlined in algorithm \ref{alg:mbrl}, which is similar to the off-policy Dyna-style architecture presented by \cite{DBLP:journals/corr/abs-1806-01825}, \cite{janner2019trust} and \cite{Kaiser2020Model}. The agent is trained for $N$ environment interaction steps, which may consist of multiple episodes. As the agent interacts with the environment, the transition tuples $(s,a,r,s')$ are added to an experience database $\mathcal D$. The agent continuously updates the value function and improves the policy by replaying experience from the database $\mathcal D$ in a model-free fashion outlined in section~\ref{sec:MFRL}. The model is updated and used only every $P$ steps i.e., the entire MBRL training is divided into $N/P$ phases. At the end of every phase i.e., every $P$ steps, the model is supervised-trained using the entire data collected so far. The rollout length $K$ is adjusted and the model is used for performing $M$ rollouts, each rooted at a different uniformly sampled experienced state. The rollouts are performed using the current policy and may include exploratory actions. The synthetic data thus collected is recorded in an experience database $\mathcal D'$ which is used to update the value function and improve the policy in a model-free fashion outlined in section~\ref{sec:MFRL}.

\algnewcommand{\Or}{\textbf{ or }}
\algnewcommand{\And}{\textbf{ and }}
\algnewcommand{\Not}{\textbf{not }}

\begin{algorithm}[t]
    \caption{Dyna Style Model-Based Reinforcement Learning}
    \begin{algorithmic}[1]
        \State Initialize policy $\pi_\theta$, predictive model $\mathcal M_\phi$, experience buffer $\mathcal D$, experience buffer $\mathcal D'$
        \For{$t = 1...N$}
            \State Act in environment using $\pi_\theta$; add experience to $\mathcal D$
            \State Train $\pi_\theta$ from $\mathcal D$ for G gradient updates
            \If{$i$ mod $P = 0$}
                \State Train model $\mathcal M_\phi$ from $\mathcal D$ using supervised learning
                \State Adjust rollout length $K$ using metareasoning
                \State Empty $\mathcal D'$
                \For{$M$ rollouts}
                    \State Sample state $s$ uniformly from $\mathcal D$
                    \State Perform $K$ steps of model rollout from $s$ using policy $\pi_\theta$; add experiences to $\mathcal D'$
                    \State Train $\pi_\theta$ from $\mathcal D'$ for G' gradient updates
                \EndFor
            \EndIf
        \EndFor
    \end{algorithmic}
    \label{alg:mbrl}
\end{algorithm}

We frame the task of optimizing the quality of the final policy learned by MBRL as a meta-level sequential decision-making process, specifically, as an MDP with the following characteristics.

\vspace{1em}

\noindent \textbf{Task Horizon:} An entire duration of MBRL training corresponds to one meta-level episode consisting of $N/P$ steps. The metareasoner adjusts the rollout length every $P$ steps during the training (line 7 in algorithm \ref{alg:mbrl}).

\vspace{1em}

\noindent \textbf{Action Space:} The meta-level action space consists of three actions:
\begin{itemize}
    \item \textsc{Up}: \quad \ \  $K \gets \text{ceil}(1.5K)$ if $K >0$;\  otherwise 1.
    \item \textsc{Down}: $K \gets \text{floor}(K / 2)$ if $K>1$; \ otherwise 0.
    \item \textsc{NoOp}: No change
\end{itemize}
Despite the simplicity of this action space, it allows rapid changes to the rollout length due to exponential effects of the actions. The \textsc{Down} action is more aggressive than the \textsc{Up} action to allow the meta-level policy to move towards a conservative rollout length rapidly. Another benefit is that a random walk with this action space makes the rollout length hover close to zero. Consequently, a meta-level policy that deliberately chooses higher rollout lengths, and performs well, would clearly demonstrate a need to use the model. 

\vspace{1em}

\noindent \textbf{State Space:} The meta-level state features are:
\begin{itemize}
    \item Time: Remaining environment interactions $N-t$.
    \item Current rollout length $K$.
    \item Quality: The average return ($J$-value) under the current policy.
    \item Model-error in predicting episode-returns: The average difference between the return predicted by the model under the current policy and the return observed in the environment under the same policy, when starting from the same initial state.
    Since this is not an absolute difference, the sign of the error reflects whether the model is optimistic or pessimistic.
    \item Model-error in predicting episode-lengths: The average difference between the episode length predicted by the model under the current policy and the episode length observed in the environment under the same policy, when starting from the same initial state.
\end{itemize}
The bottom three features are computed by taking average over the episodes that take place during the latest phase.

\vspace{1em}

\noindent \textbf{Reward Function:} The reward for the meta-level policy is the change in the average return ($J' - J$) of the current MBRL policy since the latest action. There is no discounting.

With this reward structure, the cumulative return becomes $J_1 + J_2 - J_1 + ... + J_{N/P} - J_{N/P-1} = J_{N/P}$. In other words, this reward structure incentivizes the meta-level agent to maximize the quality of the final policy learned by the MBRL agent.

\vspace{1em}

\noindent \textbf{Transition Function:} The effects of actions on the state features of the meta-MDP are not available explicitly.

\vspace{1em}

As the meta-level transition model is explicitly unknown, solving the meta-MDP requires a model-free approach, such as model-free reinforcement learning. In our approach, the meta-level RL agent, or the metareasoner, is trained using model-free deep reinforcement learning on instances of an MBRL agent solving a simplified simulation of the target real world environment, so that the trained metareasoner can be transferred to the real world to achieve our objective of helping the MBRL agent learn in real-world environments given a fixed budget of environment interactions.

\section{Related work}

While there has been substantial work on developing model-learning and model-usage techniques for reliably predicting long trajectories~\citep{DBLP:journals/corr/abs-2006-16712}, little work has gone into adapting the rollout length in a principled manner. 

\cite{DBLP:journals/corr/abs-1806-01825} study the effect of planning shape (number of rollouts and rollout length) on Dyna-style MBRL and observe that one-step Dyna offers little benefit over model-free methods in high dimensional domains, concluding that the rollouts must be longer for the model to generate unfamiliar experience.

\cite{nguyen2018improving} argue that transitions synthesized from the learned model are more useful in the beginning of the training when the model-free value estimates are only beginning to converge, and less useful once real data is abundant enough that it provides more accurate training targets for the agent. They suggest that an ideal rollout strategy would roll out more steps in the beginning, and less at the end. However, their efforts to adaptively truncate rollouts based on estimates of the model's and the value function's uncertainty have met little success.

\cite{janner2019trust} derive bounds on policy improvement using the model, based on the choice of the rollout length and the model's ability to generalize beyond its training distribution. Their analysis suggests that it is safe to increase the rollout length linearly, as the model becomes more accurate over the course of the training, to obtain maximal sample efficiency while still guaranteeing monotonic improvement. While this approach is effective, it does not consider long term effects of the modifying the rollout length.

To our knowledge, ours is the first approach that performs a sequence of adjustments to the rollout length based on the model's empirical error, the performance of the agent, and the remaining budget of environment interactions, to optimize the quality of the final policy in a decision theoretic manner. Our formulation of the problem as a meta-level MDP and use of deep reinforcement learning to solve it is inspired from prior literature~\citep{biedenkapp2020dynamic, bhatia2022tuning}.

\section{Experiments}

We experiment with \textsc{Dqn}~\citep{mnih2015human} model-free RL algorithm as a metareasoner for adjusting the rollout length in \textsc{Dyna-Dqn}~\citep{DBLP:journals/corr/abs-1806-01825} model-based RL algorithm on popular RL environments \textsc{MountainCar}~\citep{moore1990efficient,singh1996reinforcement} and \textsc{Acrobot}~\citep{sutton1996generalization,JMLR:v16:geramifard15a}. 

For each environment, the \textsc{Dqn} metareasoner is trained for 2000 meta-level episodes -- each corresponding to one training run of \textsc{Dyna-Dqn} consisting 150k steps and 120k steps on \textsc{MountainCar} and \textsc{Acrobot} environments respectively. The rollout length is capped at 32, and adjusted every 10k steps, so that the meta-level task horizon is 15 steps and 12 steps respectively. The metareasoner's score for each training run is calculated by evaluating the final policy of the \textsc{Dyna-Dqn} agent for that run, without exploration, averaged over 100 episodes of the environment. The overall score of the trained metareasoner is calculated by taking its average score over 100 training runs on test environments, which are constructed by perturbing the parameters of the original environments in order to capture the gap between the real world and the simulation. This is done to test the metareasoner’s ability to transfer to the real world, which is our ultimate objective.

Further details and hyperparameters of $\textsc{Dqn}$, $\textsc{Dyna-Dqn}$ and the modified RL environments are in the appendix.

We compare our approach $\textsc{Meta}$ to various rollout length schedules that have been suggested in prior literature: i) $K=0$ (i.e., Model-free $\textsc{Dqn}$)  ii) static rollout lengths $K=16$ and $K=32$, iii) $\textsc{Dec}$: linearly decrease $K = 32 \rightarrow 0$ over the course of the training, iv) $\textsc{Inc}$: linearly increase $K = 0 \rightarrow 32$, and v)  $\textsc{Inc-Dec}$: linearly increase then decrease $K= 0 \rightarrow 32 \rightarrow 0$ over the two halves of the training.

\section{Results and Discussion}

\begin{table}[t]
    \centering
    \small \begin{tabular}{| c | c | c | c | c | c | c | c |}
        \hline
        Environment & $K=0$ & $K=16$ & $K=32$ & $\textsc{Dec}$ & $\textsc{Inc}$ & $\textsc{Inc-Dec}$ & $\textsc{Meta}$ \\
        \hline
        $\textsc{MountainCar}$ & $-188.9$ &  $-166.0$ &  $-170.0$ & $-166.7$ & $-168.0$ & $-165.8$ & $\mathbf{-160.0}$ \\
         & $(\pm 0.7)$ &  $(\pm1.4)$ &  $(\pm1.4)$ & $(\pm1.3)$ & $(\pm 1.4)$ & $(\pm1.4)$ & $(\pm 1.3)$ \\
        $\textsc{Acrobot}$ & $-151.34$ &  $-142.8$ &  $-148.1$ & $-148.0$ & $-144.6$ & $-140.5$ & $\mathbf{-130.7}$ \\
        & $(\pm 3.4)$ &  $(\pm4.0)$ &  $(\pm3.9)$ & $(\pm3.9)$ & $(\pm 4.1)$ & $(\pm 3.7)$ & $(\pm 3.5)$ \\
        \hline
    \end{tabular}
    \vspace{0.5em}
    \caption{Mean score ($\pm$ standard error) of each approach across 100 training runs of \textsc{Dyna-Dqn} on the modified (i.e., test) \textsc{MountainCar} and \textsc{Acrobot} environments.}
    \label{table:results}
\end{table}

\begin{figure}[t]
    \centering
    \begin{subfigure}{0.24\textwidth}
        \centering
        \includegraphics[width=\linewidth]{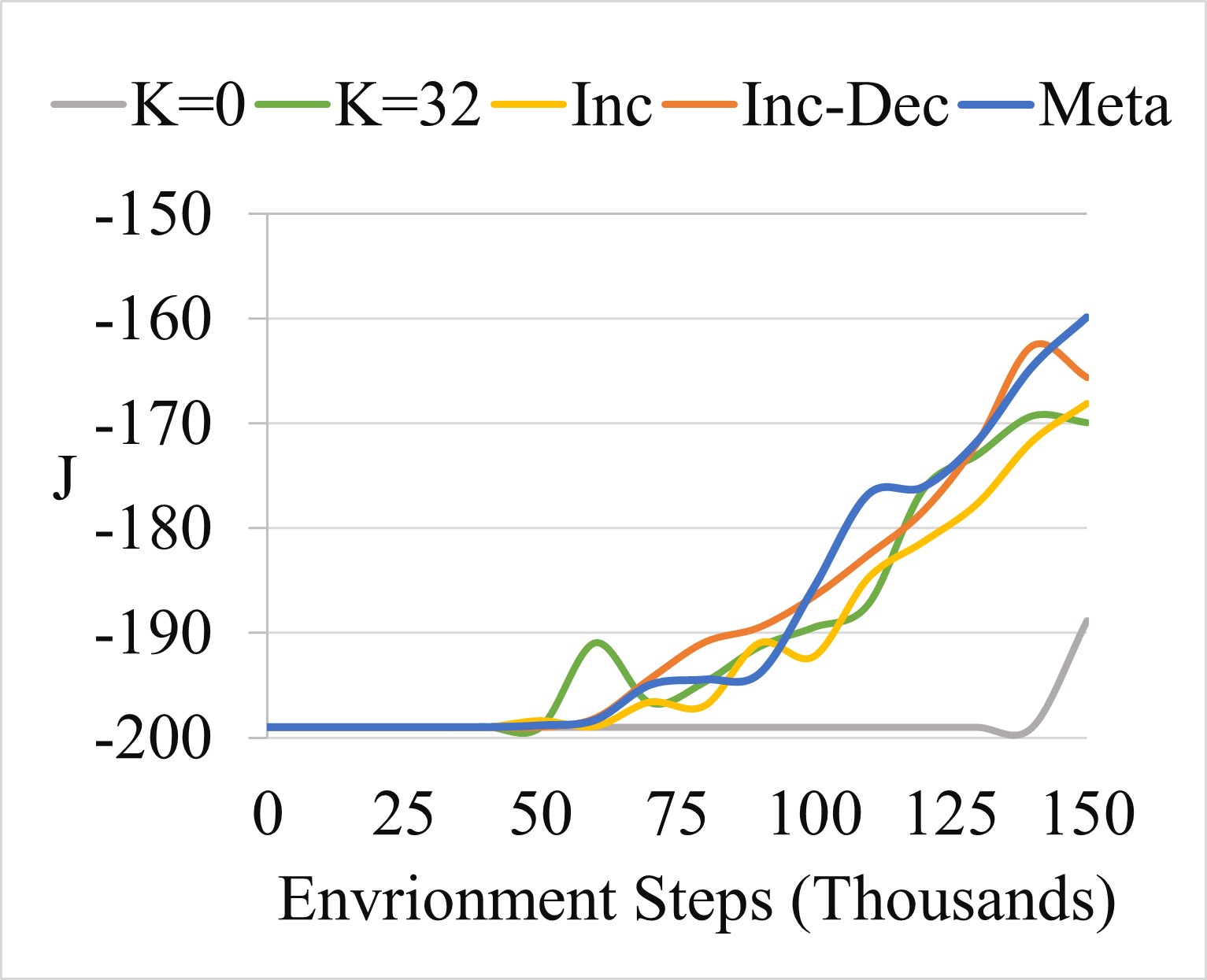}
    \end{subfigure}
    \begin{subfigure}{0.24\textwidth}
        \centering
        \includegraphics[width=\linewidth]{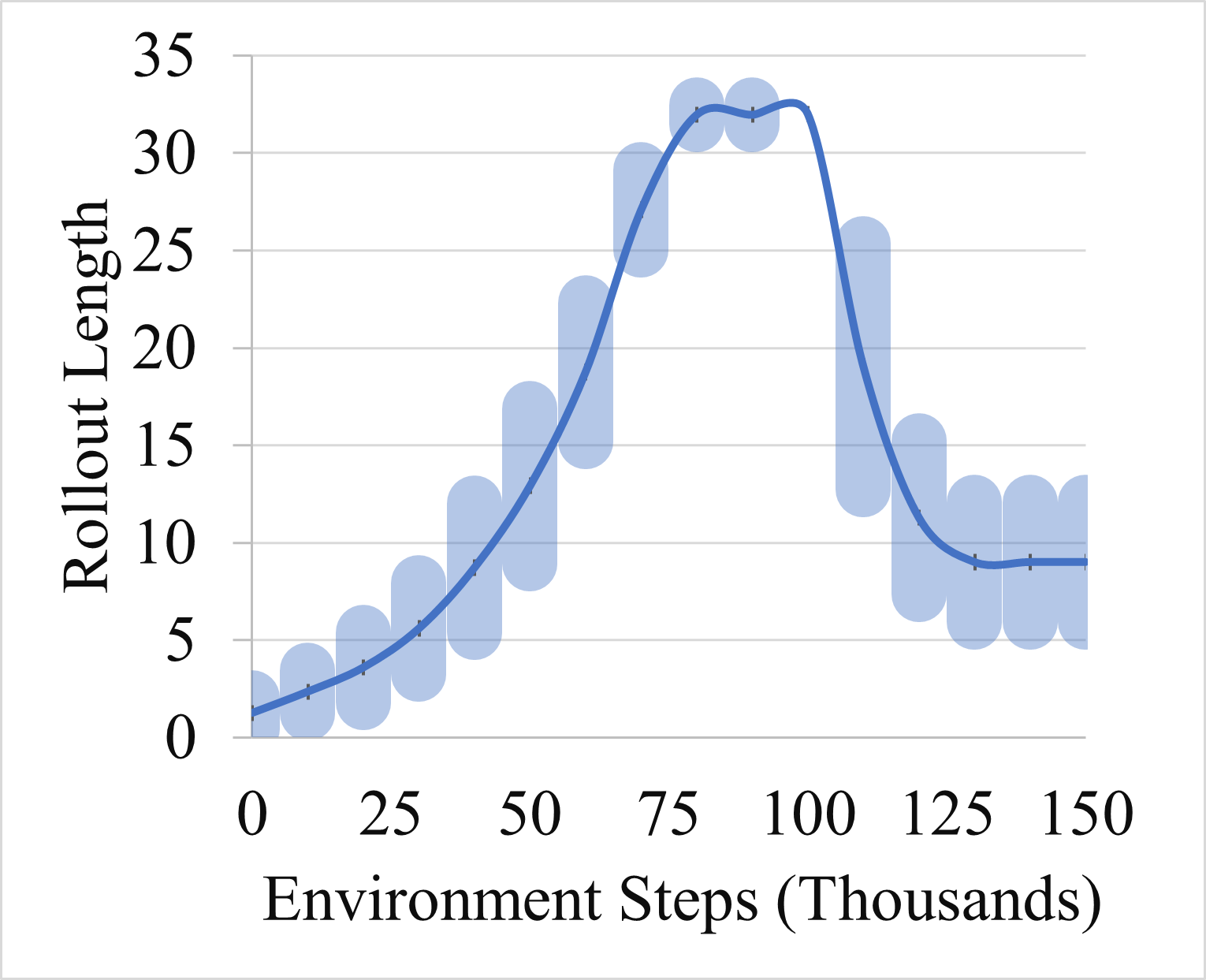}
    \end{subfigure}
    \begin{subfigure}{0.24\textwidth}
        \centering
        \includegraphics[width=\linewidth]{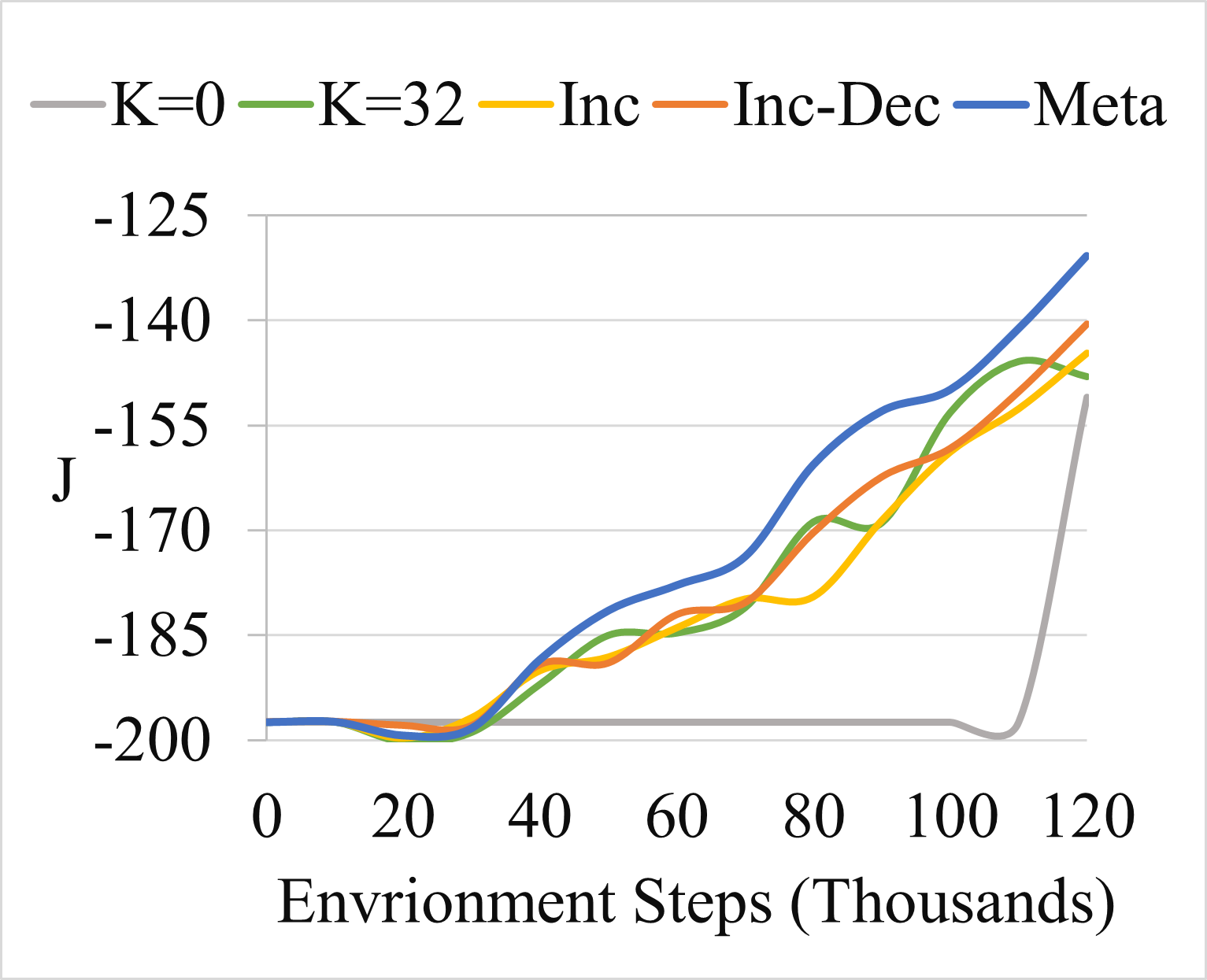}
    \end{subfigure}
    \begin{subfigure}{0.24\textwidth}
        \centering
        \includegraphics[width=\linewidth]{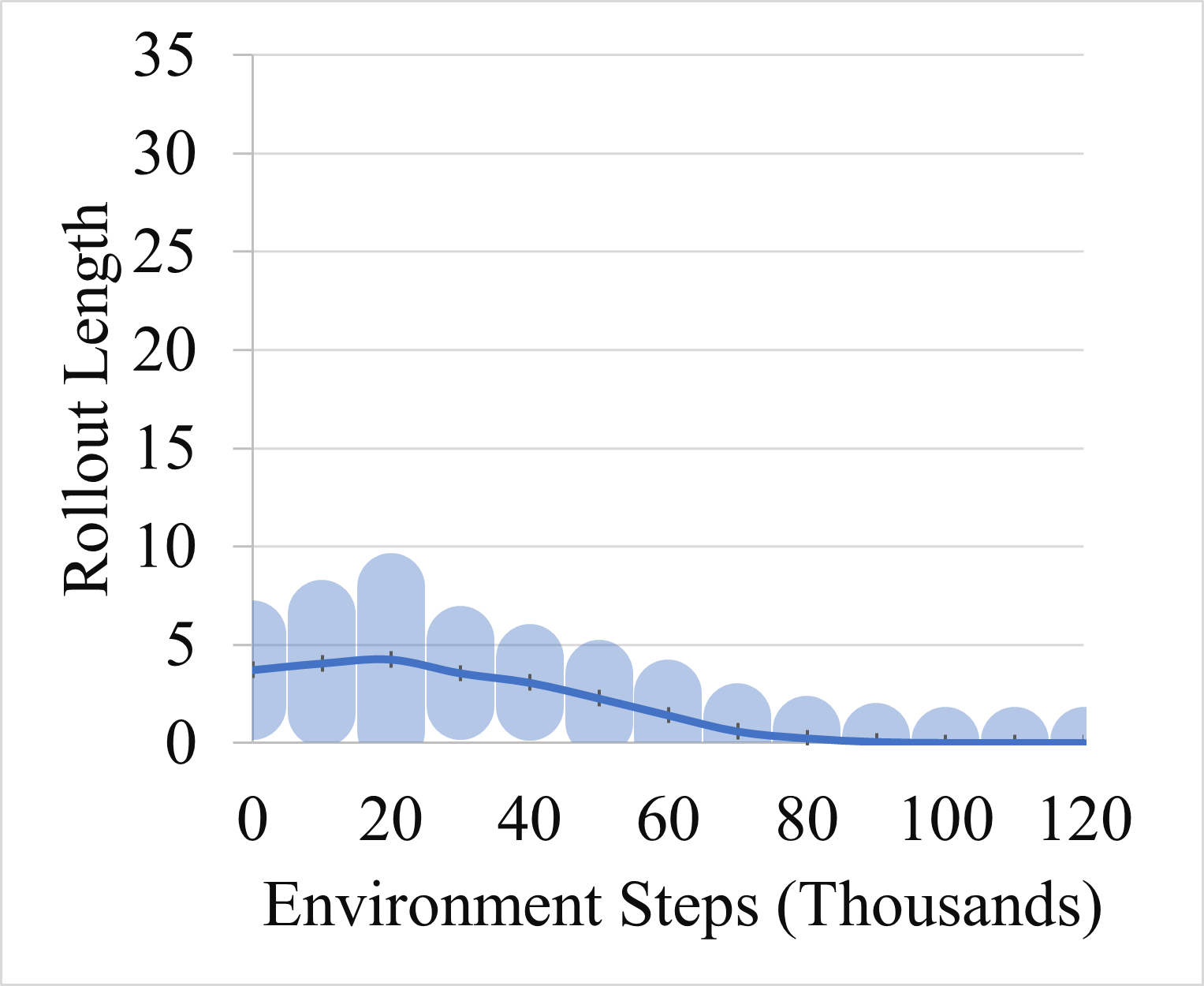}
    \end{subfigure}
    \caption{(a) Mean learning curve across the training runs of \textsc{Dyna-Dqn} on the modified \textsc{MountainCar} environment for selected rollout adjustment approaches; (b) Mean ($\pm$ standard deviation) rollout length schedule learned by the metareasoner on the modified \textsc{MountainCar} environment; (c) Mean learning curve of \textsc{Dyna-Dqn} on the modified \textsc{Acrobot} environment for selected rollout adjustment approaches; (d) Mean ($\pm$ standard deviation) rollout length schedule learned by the metareasoner on the modified \textsc{Acrobot} environment.}
    \label{fig:results}
\end{figure}

Table \ref{table:results} shows the mean score (and standard error) of each approach across the 100 training runs of \textsc{Dyna-Dqn} on the modified (i.e., test) RL environments. On both environments, the metareasoning approach demonstrates better mean score than that of the baseline approaches. Figures \ref{fig:results}(a) and \ref{fig:results}(c) show the mean learning curves of \textsc{Dyna-Dqn} for selected rollout adjustment approaches. The learning curves demonstrate that the metareasoning approach to rollout adjustment leads to more stable learning curves on average than the baseline approaches.

The results on \textsc{MountainCar} show that it is a difficult environment for model-free reinforcement learning to solve within a limited budget of environment interactions. On this environment, all model-based reinforcement learning approaches achieve significantly higher scores. Among the model-based approaches,  while all heuristic rollout schedules demonstrate similar performance, the metareasoning approach outperforms every other approach. On \textsc{Acrobot}, most model based approaches lead to minor gains over the model-free baseline at the end of the budget. However, the metareasoning approach achieves significantly higher mean score. \textsc{Inc-Dec} approach performs the best among the baseline rollout schedules.

On both domains, the metareasoning approach leads to better learning curves on average than that of the baselines. On \textsc{MountainCar}, it leads to comparatively monotonic improvement on average. On \textsc{Acrobot}, the learning curve due to the metareasoning approach dominates the other learning curves at all times during the training on average.

Figure \ref{fig:results}(b) shows the rollout length chosen by the trained metareasoner on average across the training runs of \textsc{Dyna-Dqn} on the modified \textsc{MountainCar} environment at different points during the training. The rollout adjustment policy appears similar to \textsc{Inc-Dec} except the greater variance, indicating that the metareasoner's policy is more nuanced than a simple function of training steps as more factors are taken into consideration. Figure \ref{fig:results}(d) shows the rollout length chosen by the trained metareasoner on average across the training runs of \textsc{Dyna-Dqn} on the modified \textsc{Acrobot} environment at different points during the training. The metareasoner chooses lower values for the most part and ultimately switches to model-free learning. This is not surprising given that the model-free baseline is competitive with the model-based approaches on this environment.

Finally, we observe a pattern that the approaches that decrease the rollout length towards the end of the training generally perform better on both environments. This suggests that even though the model becomes more accurate as the training progresses, its net utility diminishes when real experience data becomes more abundant. However, this observation may not generalize beyond our choice of the environments, the interactions budget and the parameters of the model-based reinforcement learning algorithm.

\section{Conclusion}

In this work, we motivate the importance of choosing and adjusting the rollout length in a principled manner during training in model-based reinforcement learning given a fixed budget of environment interactions in order to optimize the quality of the final policy learned by the agent. We frame the problem as a meta-level closed-loop sequential decision-making problem such that the adjustment strategy incorporates feedback from the learning process, which includes features such as improvement in the model's accuracy as training progresses. We solve the meta-level decision problem using model-free deep reinforcement learning and demonstrate that this metareasoning approach leads to more stable learning curves and ultimately a better final policy on average as compared to certain heuristic approaches.

\section{Acknowledgments}

This work was supported by NSF grants IIS-1813490 and IIS-1954782.

\bibliographystyle{apalike}
\bibliography{bibliography}

\newpage

\section{Appendix}

\subsection{RL Environments}

We used the default implementations of \textsc{MountainCar} and \textsc{Acrobot} in \emph{ReinforcementLearning.jl}~\citep{Tian2020Reinforcement} Julia library. The parameter modifications for testing the metareasoner are as follows.

\begin{table}[h!]
    \begin{tabular}{| c | c |}
        \hline
        Environment & Modifications \\
        \hline
        \begin{tabular}{c}
            \textsc{MountainCar} \\
            \\
            \textsc{Acrobot} \\
            \\
            \\
            \\
            \\
        \end{tabular} & \begin{tabular}{c | c}
            Gravity         & 0.0025 $\rightarrow$ 0.003 \\
            Goal position   & 0.5 $\rightarrow$ -1.1 \\
            \hline
            Gravity         & 9.8 $\rightarrow$ 12.0 \\
            Link A length   & 1.0 $\rightarrow$ 1.2 \\
            Link A mass     & 1.0 $\rightarrow$ 1.2 \\
            Link B length   & 1.0 $\rightarrow$ 0.8 \\
            Link B mass     & 1.0 $\rightarrow$ 0.8
        \end{tabular} \\
        \hline
    \end{tabular}
\end{table}

\subsection{\textsc{Dyna-Dqn} Model-Based Reinforcement Learning}

This subsection describes the details of the \textsc{Dyna-Dqn} algorithm \ref{alg:mbrl}.  

\noindent \textbf{Model Learning:} Given a state and an action, a deterministic model $\mathcal M_\phi$ predicts the next state, reward and the log-probability whether the next state is terminal. The model has a subnetwork corresponding to each subtask, without any shared parameters. The transition (next state) subnetwork and the terminal subnetwork both have two hidden layers of size [32,16]. The reward subnetwork has two hidden layers of size [64,32]. ReLU activation function is used in the hidden layers, while the final layers are linear. Layer normalization~\citep{ba2016layer} is applied before each activation. Following~\cite{janner2019trust}, the transition subnetwork first predicts the difference between the next state and the current state, and then reconstructs the next state using the difference. The transition and the reward subnetworks are trained to minimize the mean squared error, while the terminal subnetwork uses binary cross-entropy loss. Adam optimizer~\citep{DBLP:journals/corr/KingmaB14} is used with learning rate $0.001$. The minibatch size is 32. The standard practice of splitting the training data into training and validation sets is followed, with 20\% data reserved for validation. The training stops when the validation loss increases.

\noindent \textbf{\textsc{Dqn} Network:} The \textsc{Dqn} network $q_\theta$ consists of two hidden layers of size [64, 32], and uses ReLU activation function. It is trained on real and model-synthesized experience using Adam optimizer with learning rate 0.0001. The minibatch size is 32. The network parameters $\theta$ are copied to $\theta'$ every 2000 gradient updates.

\noindent \textbf{Acting, Learning and Planning Loop:} Total environment steps $N$ is 150k and 120k for \textsc{MountainCar} and \textsc{Acrobot} respectively. The experience buffer $\mathcal D$ is of unlimited capacity. The rollout length is adjusted every $P$=10k steps. The total number of rollouts are such that the total number of rollout steps match the total number of environment steps, i.e., $M=P/K$. The rollouts are truncated at $K$ steps, or if the model transitions to a terminal state, whichever happens earlier. Both the acting policy and the rollout policy includes $\epsilon=0.1$ exploration. For the acting policy, $\epsilon=1$ for the initial 10k steps and is linearly annealed to $0.1$ over the next 10k steps. For each actual or synthetic experience, $G=G'=1$ gradient update is performed. The double-Q-learning update rule is used to reduce maximization bias~\citep{van2016deep, hasselt2010double}. The discount factor $\gamma$ is $0.99$.

\subsection{\textsc{Dqn} Metareasoning}

The \textsc{Dqn} metareasoner is trained on 2000 meta-level episodes with a different seed used for each training run of \textsc{Dyna-Dqn}. $\epsilon$-greedy exploration is used with $\epsilon=1$ for the first 25 episodes and linearly annealed to $\epsilon=0.15$ over the next 25 episodes. A discount factor of 0.99 is used. The network uses two hidden layers of size [64, 32] with ReLU activation units. 10 gradient updates are performed for each meta-level experience using the double-Q-learning update rule, with minibatch size 32, and Adam learning rate 0.0001. The network parameters are copied to the target network every 10 meta-level episodes.

\end{document}